\documentclass[letterpaper, 10 pt, conference]{ieeeconf}  % Comment this line out if you need a4paper

\IEEEoverridecommandlockouts                              % This command is only needed if 
\usepackage{times}
\usepackage[pdftex]{graphicx}
\usepackage{subfigure}
\usepackage{amsmath,amssymb,amsopn,amstext,amsfonts}
\usepackage{cancel}
\usepackage[space]{cite}
\usepackage{pdfsync}
\usepackage{balance}
\usepackage{color}
\usepackage{mathtools}
\usepackage{bm}

\usepackage{diagbox}
\usepackage{float}
\usepackage{epstopdf}
\usepackage{pifont}
\usepackage{fixltx2e}
\usepackage{amsmath}
\usepackage{multirow}
\usepackage{url}
\usepackage{verbatim}
\usepackage[linkcolor=black,citecolor=black,urlcolor=black,colorlinks=true]{hyperref}
\usepackage{soul,xcolor}
\usepackage[linesnumbered,ruled,vlined]{algorithm2e}

\graphicspath{{../img/}}
\DeclareGraphicsExtensions{.png,.jpg,.eps,.pdf,.jpeg}

\title{\LARGE \bf
Robust Real-time UAV Replanning Using \\ Guided Gradient-based Optimization and Topological Paths
}

\author{Boyu Zhou, Fei Gao, Jie Pan, and Shaojie Shen%
\thanks{This work was supported by HKUST-DJI Joint Innovation Laboratory and HKUST institutional fund. All authors are with the Department of Electronic and Computer Engineering, Hong Kong University of Science and Technology, Hong Kong, China. {\tt\footnotesize $\{$bzhouai, fgaoaa, jpanai, eeshaojie$\}$@ust.hk}}%
}

\begin{document}

\maketitle
\thispagestyle{empty}
\pagestyle{empty}

%%%%%%%%%%%%%%%%%%%%%%%%%%%%%%%%%%%%%%%%%%%%%%%%%%%%%%%%%%%%%%%%%%%%%%%%%%%%%%%%
\begin{abstract}

Gradient-based trajectory optimization (GTO) has gained wide popularity for quadrotor trajectory replanning. 
However, it suffers from local minima, which is not only fatal to safety but also unfavorable for smooth navigation.
In this paper, we propose a replanning method based on GTO addressing this issue systematically.
A path-guided optimization (PGO) approach is devised to tackle infeasible local minima, which improves the replanning success rate significantly.    
A topological path searching algorithm is developed to capture a collection of distinct useful paths in 3-D environments, each of which then guides an independent trajectory optimization.
It activates a more comprehensive exploration of the solution space and output superior replanned trajectories. 
Benchmark evaluation shows that our method outplays state-of-the-art methods regarding replanning success rate and optimality.
Challenging experiments of aggressive autonomous flight are presented to demonstrate the robustness of our method.
We will release our implementation as an open-source package\footnote{The open-source code will be released at \url{https://github.com/HKUST-Aerial-Robotics/Fast-Planner} and \url{https://github.com/HKUST-Aerial-Robotics/TopoTraj}}.

\end{abstract}

%%%%%%%%%%%%%%%%%%%%%%%%%%%%%%%%%%%%%%%%%%%%%%%%%%%%%%%%%%%%%%%%%%%%%%%%%%%%%%%%
\section{Introduction}
Unmanned aerial vehicles (UAVs) are gaining popularity for many applications such as autonomous inspection, transportation, and photography, in which the UAV is required to navigate along a global reference trajectory to finish its missions. 
To ensure safety, trajectory replanning is of vital importance to cope with previously unknown obstacles.

The problem of trajectory replanning has been investigated actively.
Recent works \cite{oleynikova2016continuous, fei2017iros, usenko2017real, zhou2019robust, gao2019teach} reveal that \textbf{g}radient-based \textbf{t}rajectory \textbf{o}ptimization (GTO), which typically formulates trajectory replanning as a non-linear optimization problem that trades off smoothness, safety, and dynamically feasibility, is particularly effective for this problem. 
It is widely applied thanks to its convenience to deform an infeasible trajectory segment into a feasible one, with very high efficiency and low memory requirement.
% since it can be convinently applied to deform a infeasible trajectory segment into a feasible one with very high efficiency and low memory requirement.  

Despite its advantages, GTO-based replanning is cursed by the issue of local minima. 
The issue arises from the collision cost of the optimization, which should be evaluated according to the structure of the environment. 
Since there are multiple safe and unsafe regions, this cost is non-convex by nature.
% Combined with the smoothness and dynamic feasibility cost, it leads to a non-convex optimization problem.  
% the non-convex Euclidean signed distance field (ESDF). 
This issue could cause fatal crashes since it frequently makes the trajectory get stuck in infeasible local minima and results in replanning failure. 
Besides, it also leads to the lack of optimality guarantee, as only a small fraction of solution space around the initial trajectory is searched. 
Consequently, the so-called local optimal trajectory is usually unsatisfactory for smooth and safe flights. 
Researchers have realized this critical issue and employed strategies like random restart \cite{oleynikova2016continuous} and iterative refinement \cite{gao2019teach} to relieve it.
Nonetheless, the problem is not resolved essentially and prohibits GTO to be applied to more challenging scenarios such as aggressive flight.

\begin{figure}[t]
	\begin{center}          
		\subfigure
		{\includegraphics[width=0.49\columnwidth]{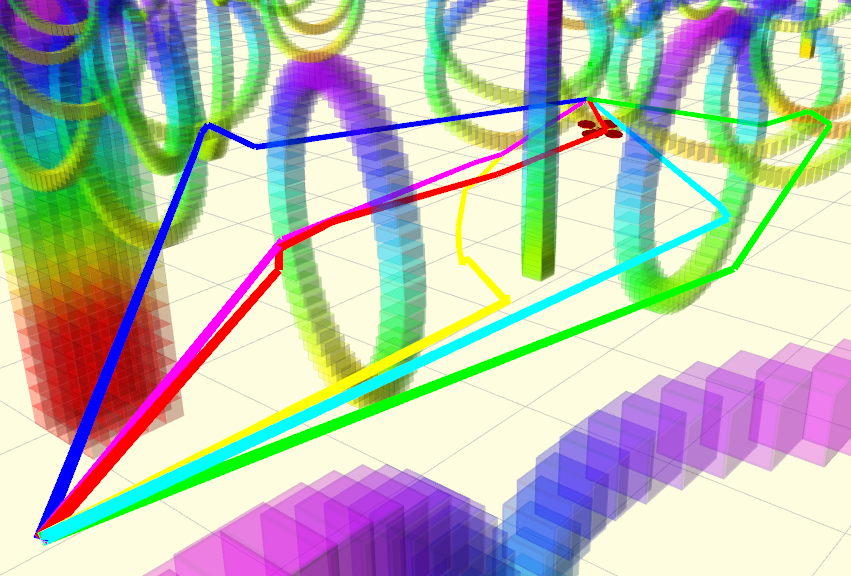}}       
		\subfigure
      {\includegraphics[width=0.49\columnwidth]{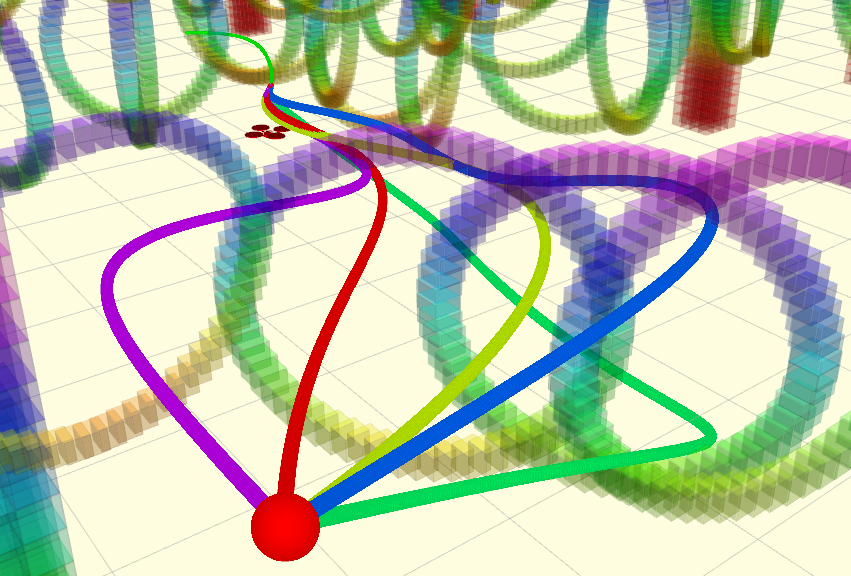}}     
      \vspace{-0.5cm}
	\end{center}
   \caption{\label{fig:topo_traj} A set of topologically distinct paths (left). Multiple trajectories generated with the guidance of different paths (right). 
   Video demonstrating the proposed replanning method in aggressive flights is available at \url{https://www.youtube.com/watch?v=YcEaFTjs-a0}.
   }
   \vspace{-0.9cm}
\end{figure}

In this paper, the local minima problem is addressed systematically by a new GTO-based replanning method, which comprised of a \textbf{p}ath-\textbf{g}uided \textbf{o}ptimization (PGO), 
an efficient algorithm to discover topologically distinct paths, and the parallel trajectory optimization guided by the paths (as depicted in Fig.\ref{fig:topo_traj}).
Firstly, we answer the question of how infeasible local minima of GTO can be reliably prevented. 
The typical reasons for infeasible local minima are investigated. 
Based on them, we propose PGO in which a geometric path is utilized to guide the optimization effectively so that the success of replanning is guaranteed. 
Secondly, we answer how optimality of the replanning can be improved considerably, with only minor computational overhead. 
We propose an efficient sampling-based topological path searching approach to extract a comprehensive set of paths that capture the structure of the environment. 
With the guidance of several carefully selected paths, PGO is invoked to explore the solution space more thoroughly.
It consistently yields better replanning than previous methods, while the total computation time is comparable.
%  and generates a rich set of local optimal trajectories, which consistently yields better replanning than previous method.
% effectively explore the essential part of search space in parallel and produce a rich set of local optimums,   

We conduct extensive benchmark comparisons with state-of-the-art methods and challenging real-world experiments to validate our proposed replanning method. 
Results show that it is superior to previous ones in terms of significantly higher success rate and stronger optimality guarantee, with only slightly longer computational time induced by the topological path searching. 
The contributions of this paper are summarized as follows:

1) A robust optimization approach for real-time trajectory replanning named PGO, to boost the success rate of the replanning.

2) An efficient topological path searching algorithm, and its integration with the proposed PGO, to search the solution space more thoroughly and yield better replanning.

3) Benchmark comparisons and real-world experiments that validate the performance of our proposed method. The source code of our implementation will be released.

% We review related literature in Sect.\ref{sec:related} and present our PGO approach in Sect. \ref{sec:pgo}. 
% The topological path searching and its integration with PGO are detailed in Sect. \ref{sec:topo_path} and \ref{sec:topo_traj}.
% Results of benchmark and both indoor and outdoor experiments are given in Sect.\ref{sec: result}.
% The paper is concluded in Sect.\ref{sec:conclude}.

\section{Related Work}
\label{sec:related}

\subsection{Gradient-based Trajectory Optimization}

GTO is one of the major trajectory generation approaches, which formulates the problem as a non-linear optimization that minimizes an objective function. 
Interest in this method was revived recently by\cite{zucker2013chomp}, which generates discrete-time trajectories by minimizing its smoothness and collision costs using covariant gradient descent. 
\cite{kalakrishnan2011stomp} has a similar formulation, but solves the problem by sampling neighboring candidates iteratively. 
The stochastic sampling strategy partially overcomes the local minima issue but is computationally intensive. 
\cite{oleynikova2016continuous} extended the method to continuous-time polynomial trajectories to avoid differential errors. 
It also does random trajectory perturbation and optimization restart for a higher success rate. However, the improvement is insignificant.  
\cite{fei2017iros} improved the success rate by providing a high-quality initial path, which is found by an informed sampling-based path searching.
However, it only applies to low-speed flight.
In \cite{usenko2017real}, the trajectory is parameterized as a uniform B-spline. 
It showed that the continuity and locality properties of B-spline are particularly useful for trajectory replanning.
\cite{zhou2019robust} further exploited the convex hull property of B-spline and improve the optimization efficiency and robustness by a large margin.
However, given a poor initial trajectory in complex environments, this method still suffers.
As a result, \cite{gao2019teach} adopts an iterative post-process to improve the practical success rate of \cite{zhou2019robust}.
By far, local minima still remains a challenge, since no method copes with it essentially.
In this paper, we propose PGO, which incorporates a geometric path in the optimization. 
As the path effectually guides the optimization to escape from infeasible local minima, the planning success rate is guaranteed. 
Moreover, multiple distinct paths produced by the topological path searching are integrated with the PGO to seek for plentiful locally optimal solutions, which ensures higher trajectory quality.

\subsection{Topological Path Planning}

% The construction of the roadmap and filtering of equivalent paths are performed simultaneously. 
There have been works utilizing the idea of topologically distinct paths for planning, in which paths belonging to different homotopy (homology) \cite{schmitzberger2002capture,rosmann2015planning,rosmann2017integrated, bhattacharya2010search, bhattacharya2012topological}
or visibility deformation \cite{jaillet2008path} classes are sought. 
\cite{schmitzberger2002capture} constructs a variant of probabilistic roadmap (PRM) to capture homotopy classes, in which path searching and redundant path filtering are conducted simultaneously. 
In contrast, \cite{rosmann2015planning,rosmann2017integrated} firstly creates a PRM or Voronoi diagram, 
after which a homology equivalence relation based on complex analysis\cite{bhattacharya2010search} is adopted to filter out redundant paths. 
These methods only apply to 2-D scenarios. 
To seek for 3-D homology classes, \cite{bhattacharya2012topological} exploited the theory of electromagnetism and propose a 3-D homology equivalence relation. 
However, it requires occupied space to be decomposed into "genus 1" obstacles, which is usually impractical.
Besides, capturing only homotopy classes in 3-D space is insufficient to encode the set of useful paths, as indicated in \cite{jaillet2008path}, since 3-D homotopic paths may be too hard to deform into each other.
To this end, \cite{jaillet2008path} leverages a visibility deformation roadmap to search for a richer set of useful paths. 
\cite{oleynikova2018sparse, blochliger2018topomap} convert maps built from SLAM systems into sparse graphs representing the topological structure of the environments.
\cite{jaillet2008path, oleynikova2018sparse, blochliger2018topomap} focus on global offline planning and are too time-consuming for online usage. 
Our topological path searching is conceptually closest to \cite{jaillet2008path}, but with a reinvented algorithm for real-time performance. 

\begin{figure}[t]
	\begin{center}          
		\subfigure[]
		{\includegraphics[width=0.49\columnwidth]{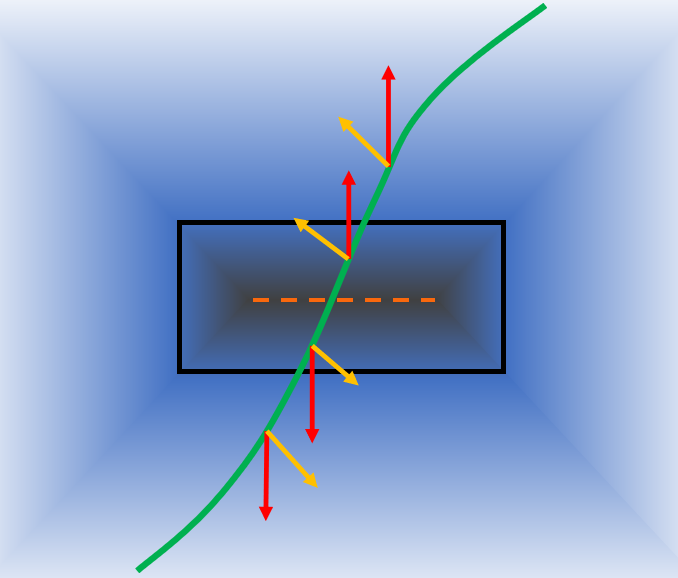}}       
		\subfigure[]
      {\includegraphics[width=0.491\columnwidth]{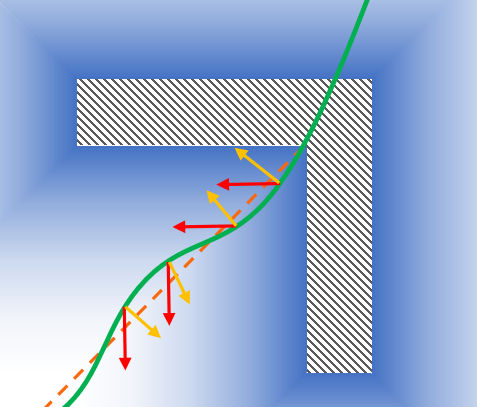}}     
      \vspace{-0.7cm}
	\end{center}
   \caption{\label{fig:failure} Typical examples of optimization failure, where adjacent parts of a trajectory are pushed in opposing directions if the trajectory crosses the "valley" (a) or "ridge" (b) of the ESDF (denoted by orange dashed lines). 
   Red arrows denote gradients of ESDF and yellow ones are gradients of the objective function. }
   \vspace{-1cm}
\end{figure}  

\section{Path-Guided Trajectory Optimization}
\label{sec:pgo}
% In this section, we focus on the design of a gradient-based optimization approach called PGO for trajectory replanning (Fig. \ref{fig:pggto}). 
% It is robust to the quality of initial path and environment complexity, and is capable of providing consistent and promising replanning.
% We start by analysing the primary cause of optimization failure in Sect. \ref{subs:fail_analy}, after which the PGO is presented accordingly in Sect. \ref{subs:pro_form}. 

\subsection{Optimization Failure Analysis} 
\label{subs:fail_analy}
Previous work\cite{schulman2014motion} showed that failure of GTO is relevant to unfavorable initialization, i.e., initial paths that pass through obstacles in certain ways usually get stuck.
Underlying reason for this phenomenon is illustrated in Fig.\ref{fig:failure}.
Typical GTO methods incorporate the gradients of a Euclidean signed distance field (ESDF) in a collision cost to push the trajectory out of obstacles.
This cost is combined with the smoothness and dynamic feasibility cost to form an objective function, whose gradients iteratively deform the trajectory into a smooth and safe one.
Yet there are some "valleys" or "ridges" in the ESDF, around which the gradients differ greatly. 
Consequently, if a trajectory is in collision and crosses such regions, the gradients of ESDF will change abruptly at some points.
This can result in gradients of the objective function pushing different parts of the trajectory in opposing directions and fails the optimization.

Normally, such "valleys" and "ridges", which corresponds to the space that has an identical distance to the surfaces of nearby obstacles, are difficult to avoid, especially in complex environments. 
Therefore, optimization depending solely on the ESDF fails inevitably at times. 
To solve the problem, it is essential to introduce extra information that can produce an objective function whose gradients consistently deform the trajectory to the free space.
% rely on the gradients of ESDF to push the trajectory out of obstacles. 
% The gradients of ESDF are incorporated in the collision cost, which is combined with the smoothness and dynamic feasibility cost to form a objective function.
% When the initial path is collision-free, gradients applied on different path segments are largely consistent (Fig.). 
% Hence, the trajectory can be pushed away from obstacles easily.
% On the contrary, if the path passes through the obstacles in certain ways (Fig.), the gradients may push different parts in in-consistent directions, and thus lock the trajectory in an infeasible local minima.
% gradients that are able to consistently deform the trajectory to the free space.

\subsection{Problem Formulation}
\label{subs:pro_form}
% We design a new two-phase gradient-based trajectory optimization method 
% We propose PGO, which consists of two different phases of optimization, to deform a segment of trajectory colliding with obstacle into a new smooth and safe one, as shown in Fig..
We propose PGO built upon our previous work \cite{zhou2019robust} that represents trajectories as B-splines for more efficient cost evaluation.
For a trajectory segment in collision, we reparameterize it as a $p_b$ degree uniform B-spline with control points $ \left\{\mathbf{Q}_{0}, \mathbf{Q}_{1}, \ldots, \mathbf{Q}_{N}\right\} $.

\begin{figure}[t]
	\begin{center}          
		\subfigure[\label{fig:pggto1}]
		{\includegraphics[width=0.41\columnwidth]{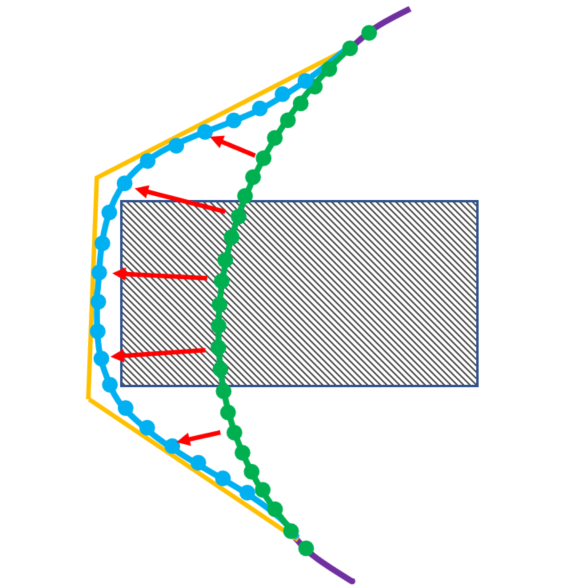}}       
		\subfigure[\label{fig:pggto2}]
		{\includegraphics[width=0.41\columnwidth]{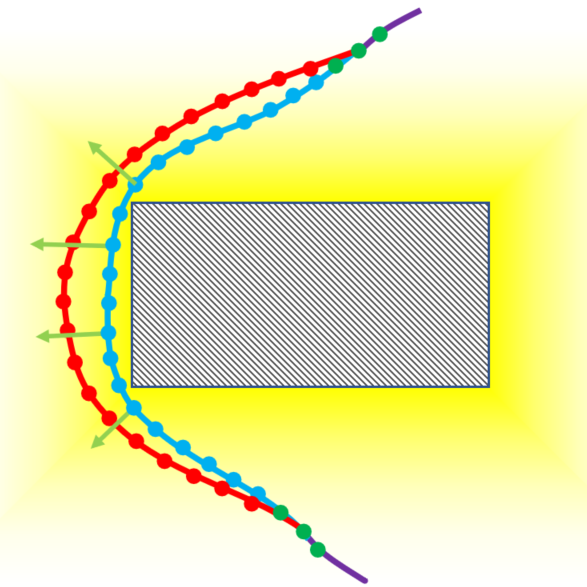}}     
      \vspace{-0.5cm}
	\end{center}
   \caption{\label{fig:pggto} The two-phases PGO approach for trajectory replanning. (a) A geometric guiding path (orange) attracts the original B-spline trajectory (green) into the free space.
   (b) The warmup trajectory is further optimized in the ESDF to improve its smoothness and clearance.}
   \vspace{-1.1cm}
\end{figure} 

% As concluded above, consistency of the gradients of ESDF along a trajectory is essential for successful optimization.
% Since the initial trajectory is in collision, attempting to push it out of obstacles directly is unfavored.
% as the source of attracting gradients 
PGO consists of two different phases. The first phase generates an intermediate \textbf{warmup trajectory}.
As concluded above, external information should be included to effectually deform the trajectory, since solely applying the ESDF could be futile.
We employ a geometric guiding path to attract the initial trajectory to the free space (depicted in Fig. \ref{fig:pggto1}) since collision-free paths are readily available from standard methods like A* and RRT*. 
In our work, the paths are provided by the sampling-based topological path searching (Sect. \ref{sec:topo_path}).
The first-phase objective function is:
\begin{equation}
   f_{p1}=\lambda_{1s} f_{s}+\lambda_{1g} f_{g}
\end{equation}
where $ f_s $ is the smoothness cost of the trajectory, while $ f_g $ is the cost to penalize the distance between the guiding path and the B-spline trajectory.
As in \cite{zhou2019robust}, $ f_s $ is designed as a elastic band cost function\footnote{Only the subset of control points $ \{ \mathbf{Q}_{p_b},\mathbf{Q}_{p_b+1}, \cdots, \mathbf{Q}_{N-p_b}\} $ is optimized 
due to the boundary state constraints of the trajectory. $ \mathbf{Q}_{p_b-2}$, $\mathbf{Q}_{p_b-1}$, $\mathbf{Q}_{N-p_b+1} $ and $ \mathbf{Q}_{N-p_b+2} $ are needed to evaluate the smoothness.} that simulates the elastic forces of a sequence of springs:
\begin{equation}
   f_{s}=\sum_{i=p_{b}-1}^{N-p_{b}+1}\left\| \mathbf{Q}_{i+1} - 2\mathbf{Q}_{i} + \mathbf{Q}_{i-1} \right\|^{2}
\end{equation}
Because the shape of a B-spline is finely controlled by its control points, we utilize this property to simplify the design of $ f_g $. 
Each control point $ \mathbf{Q}_{i} $ is assigned with an associated point $ \mathbf{G}_{i} $ on the guiding path, which is uniformly sampled along the guiding path.
Then $ f_g $ is defined as the sum of the squared Euclidean distance between these point pairs: 
% The guiding path is uniformly discretized into a sequence of points $ \mathbf{G}_{i} $, each of which is assigned to a corresponding control points $ \mathbf{Q}_{i} $. 
\begin{equation}
   f_{g}=\sum_{i=p_{b}}^{N-p_{b}} \left\| \mathbf{Q}_{i} - \mathbf{G}_{i} \right\|^{2}
\end{equation}
Notably, it is an unconstrained quadratic programming problem, so its optimal solution can be obtained in closed form.
It outputs a smooth trajectory in the vicinity of the guiding path.
Since the path is already collision-free, usually the warmup trajectory is also so.
Even though it is not completely collision-free, its major part will be attracted to the free space.
At this stage, the gradients of ESDF along the trajectory vary smoothly, and the gradients of the objective function (green arrows in Fig.\ref{fig:pggto2}) push the trajectory to the free space in consistent directions.
Hence, standard GTO methods can be utilized to improve the trajectory.

In the second phase, we adopt our previous B-spline optimization method\cite{zhou2019robust} to further refine the warmup trajectory into a smooth, safe, and dynamically feasible one, whose objective function is:
\begin{equation}
   f_{p2}=\lambda_{2s} f_{s}+\lambda_{2c} f_{c}+ \lambda_{2d}\left( f_{v} + f_{a} \right)
\end{equation}
$ f_c $ is the collision cost evaluated on the ESDF, which grows rapidly if the trajectory gets close to obstacles.
$ f_v $ and $ f_a $ penalize infeasible velocity and acceleration.
The formulations of $ f_c$, $ f_v$, and $ f_a $ are simplified based on the convex hull property of B-spline, thanks to which it suffices to constrain the control points of the B-spline to ensure safety and dynamic feasibility, without evaluating expensive line integrals.
For brevity, we refer the readers to \cite{zhou2019robust} for detailed formulations.
% $ f_c $ is formulated as the repulsive force of the obstacles acting on the control points, while 
% Note that although we can search a geometric or kinodynamic path as intitial value and do the second phase optimization as \cite{fei2017iros, zhou2019robust}, this strategy is inferior.
% Besides, searching for a kinodynamic path with boundary state constraints is expensive.
% Seaching for a high quality geometric path takes non-trivial time, while a poor path can slow down the convergence of optimization. 

Although PGO has one more step of optimization compared with previous methods, it can generate better trajectories within shorter time.
The first-phase takes only negligible time, but generate a warmup trajectory that is easier to be further refined, which improve the overall efficiency. 

\begin{figure}[t]
	\begin{center}
		\subfigure[\label{fig:short_long1}]          
		{\includegraphics[width=0.49\columnwidth]{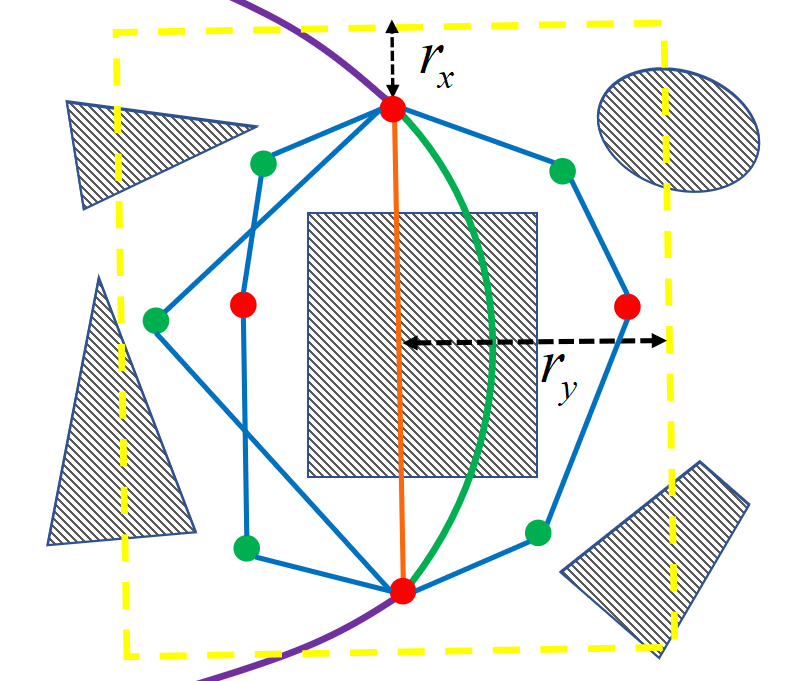}}
		\subfigure[\label{fig:short_long2}]          
      {\includegraphics[width=0.49\columnwidth]{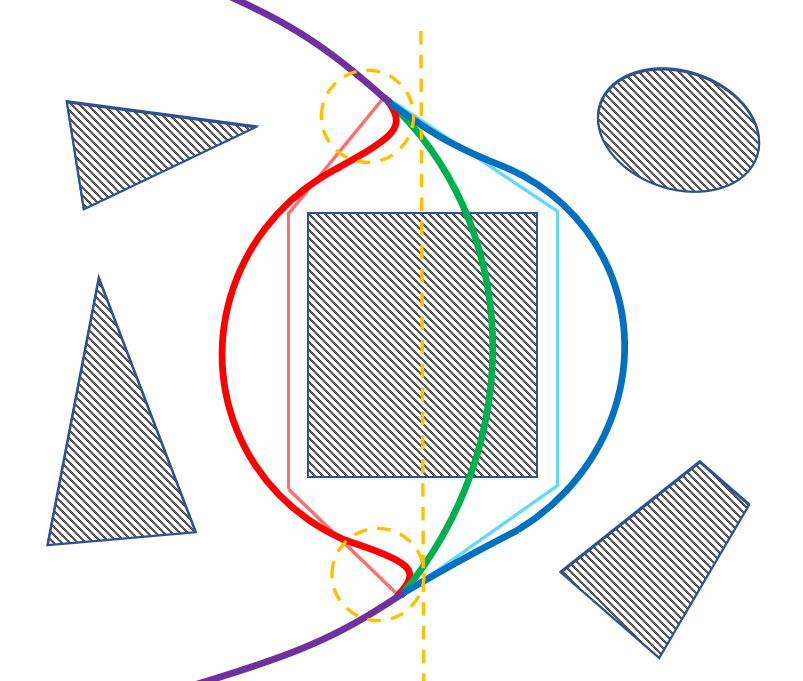}}
      \vspace{-0.75cm}
	\end{center}
   \caption{\label{fig:short_long} The proposed replanning framework.
   a) A trajectory segment within the checking horizon (green) collides with an obstacle and triggers the topological roadmap generation (Sect.\ref{subs:topo_roadmap}) within a cube.
   b) Paths are extracted, shortened and pruned to guide the PGO in parallel.
   The red path is shorter than the blue one; however, it leads to a more jerky trajectory which takes sharp turns near the start and end position. }
   \vspace{-1.5cm}
\end{figure}

\section{Topological Path Searching}
\label{sec:topo_path}
Given a geometric guiding path, our PGO method can replan a locally optimal trajectory.
However, this trajectory is not necessarily satisfactory, even with the guidance of the shortest path, as seen in Fig. \ref{fig:short_long2}.
Actually, it is difficult to determine the best guiding path, since the paths do not contain high order information (velocity, acceleration, etc.), and can not completely reflect the true motion.
Searching a kinodynamic path \cite{ding2018trajectory, liu2017iros} may suffice, but it takes excessive time to obtain a promising path with boundary state constraints at the start and end of the replanned trajectory.

\begin{figure}[t]
	\begin{center}          
		\subfigure[\label{fig:definition1} The four trajectories are equivalent under the definition of homotopy, but represent substantially different motions of a quadrotor.]
		{\includegraphics[width=0.49\columnwidth]{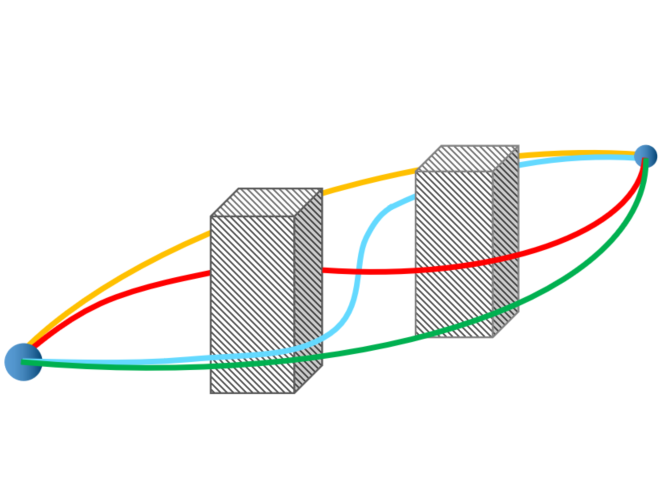}}       
		\subfigure[\label{fig:definition2} An illustration of UVD. The blue trajectory is distinct to the green one, but is equivalent to the purple one.]
		{\includegraphics[width=0.49\columnwidth]{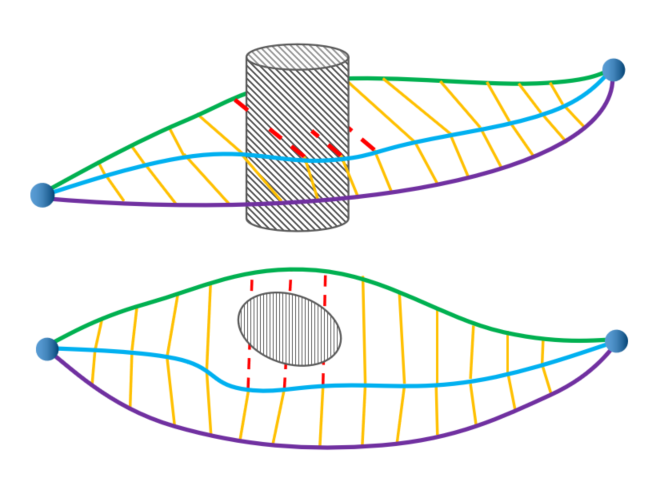}}     
	\end{center}
	\vspace{-0.3cm}
   \caption{\label{fig:definition} Topology equivalence relation.}
   \vspace{-0.99cm}
\end{figure} 

For a better solution, a variety of paths is required. 
We propose a sampling-based topological path searching to find a collection of distinct paths to guide the PGO. 
Although methods \cite{jaillet2008path,schmitzberger2002capture,oleynikova2018sparse,blochliger2018topomap} are for this problem, none of them runs in real-time in complex 3-D environments.
We redesign the algorithm carefully to solve this challenging problem in real time. 

\subsection{Topology Equivalence Relation}
\label{subs:relation}
% the four trajectories belong to the same homotopy class, but in GTO none of them can be deformed into the other.
Although the concept of homotopy is widely used, it captures insufficient useful trajectories in 3-D environments, as shown in Fig. \ref{fig:definition1}. 
\cite{jaillet2008path} proposes a more useful relation in 3-D space named visibility deformation (VD), but it is computationally expensive for equivalence checking.
Based on VD, we define \textit{uniform visibility deformation} (UVD), which also captures abundant useful trajectories, and is more efficient for equivalence checking.
% namely  that characterizes subsets of visibility deformation classes, which is more useful for our purpose.

\begin{algorithm}[h]
   \DontPrintSemicolon
   \textbf{Initialize}()\;
   $ \textbf{AddGuard}(\mathcal{G}, s), \ \textbf{AddGuard}(\mathcal{G}, g) $ \;
   \While{$ t \le t_{max} \land N_{sample} \le N_{max} $}{
      $ p_s \gets \textnormal{\textbf{Sample}}() $ \;
      $ g_{vis} \gets \textnormal{\textbf{VisibleGuards}}(\mathcal{G}, p_s) $ \;
      
      \If{$ g_{vis}.\textnormal{\textbf{size}}() == 0 $}{
         $ \textbf{AddGuard}(\mathcal{G}, p_s) $ \;
      }
      \If{$ g_{vis}.\textnormal{\textbf{size}}() == 2 $}{
         $ path_1 \gets \textnormal{\textbf{Path}}(g_{vis}[0], p_s, g_{vis}[1]) $ \;
         $ distinct \gets True $ \;
         $ \mathcal{N}_s \gets \textnormal{\textbf{SharedNeighbors}}(\mathcal{G}, g_{vis}[0], g_{vis}[1]) $ \;
         \For{$ \textnormal{\textbf{each}} \ n_s \in \mathcal{N}_s $}{
            $ path_2 \gets \textnormal{\textbf{Path}}(g_{vis}[0], n_s, g_{vis}[1]) $ \;
            \If{$ \textnormal{\textbf{Equivalent}}(path_1, path_2) $}{
               $ distinct \gets False $\;
               \If{$ \textnormal{\textbf{Len}}(path_1) < \textnormal{\textbf{Len}}(path_2) $}{
                  $ \textnormal{\textbf{Replace}}(\mathcal{G}, p_s, n_s) $\;
               }
               break\;
            }
         }
         \If{$ distinct $}{
            $ \textbf{AddConnector}(\mathcal{G}, p_s, g_{vis}[0], g_{vis}[1]) $ \;
         }
      }
   }
   \caption{Topological Roadmap \label{alg:topo_roadmap}}
   % \vspace{-0.2cm}
\end{algorithm}

\textbf{Definition 1.} Two trajectories $ \tau_1(s) $, $ \tau_2(s) $ parameterized by $ s \in [0,1] $ and satisfying $ \tau_1(0) = \tau_2(0), \tau_1(1) = \tau_2(1) $,
belong to the same \textit{uniform visibility deformation} class, if for all $ s $, line $ \overline{\tau_1(s)\tau_2(s)} $ is collision-free.

\begin{figure}[t]
	\begin{center}          
		\includegraphics[width=0.99\columnwidth]{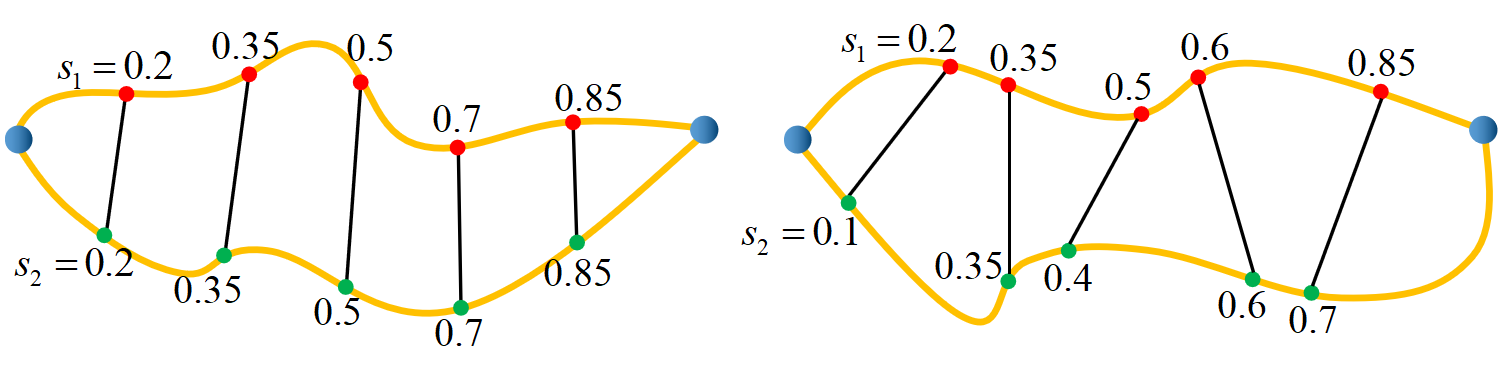}       
	\end{center}
   \vspace{-0.3cm}
   \caption{\label{fig:vduvd} Comparison between UVD (left) and VD (right).
   Each red point on one path is transformed to a point (green) on the other path.  
   Any two associated points correspond to the same parameter $ s $ in UVD, but not in VD.
   }
   \vspace{-1.9cm}
\end{figure}

Fig. \ref{fig:definition2} gives an example of three trajectories belonging to two UVD classes.
The relation between VD and UVD is depicted in Fig. \ref{fig:vduvd}.
Both of them define a continuous map between two paths $\tau_1(s)$ and $ \tau_2(s) $, in which a point on $\tau_1(s)$ is transformed to a point on $\tau_2(s)$ through a straight-line.
The major difference is that for UVD, point $ \tau_1(s_1) $ is mapped to $\tau_2(s_2)$ where $ s_1 = s_2 $, while for VD $ s_1$ does not necessarily equals $ s_2 $.
In concept, UVD is less general and characterizes subsets of VD classes.
Practically, it captures slightly more classes of distinct paths than VD, 
but is far less expensive \footnote{To test VD relation, one should compute a visibility diagram and do path searching within it\cite{jaillet2008path}, which has higher complexity than testing UVD.} for equivalence checking.

% Both the VD and UVD between two paths $\tau_1(s)$, $ \tau_2(s) $ forms a deformation surface swept out by a moving straight line, i.e., a ruled surface.
% Their major difference is that for UVD the moving line always intersects the paths at two points $\tau_1(s_1)$, $ \tau_2(s_2) $ where $ s_1 = s_2 $, while for VD $ s_1$ does not necessarily equals $ s_2 $.

To test UVD relation, one can uniformly discretize $ s \in [0,1] $ to $ s_i = i/K, i = 0, 1,..., K $ and check collision for lines $ \overline{\tau_1(s_i)\tau_2(s_i)} $.
For the piece-wise straight line paths (as in Alg. \ref{alg:topo_roadmap}, \textbf{Equivalent}()), we simply parameterize it uniformly, 
so that for any $ s $ except $ \tau(s) $ is the join points of two straight lines, $ \left\| \frac{d\tau(s)}{ds} \right\| = const  $. 
% \textbf{Equivalent}()

% \vspace{-0.5cm}
\subsection{Topological Roadmap}
\label{subs:topo_roadmap}

\begin{figure*}[t]
	\begin{center}          
		\subfigure[]
		{\includegraphics[width=0.4\columnwidth]{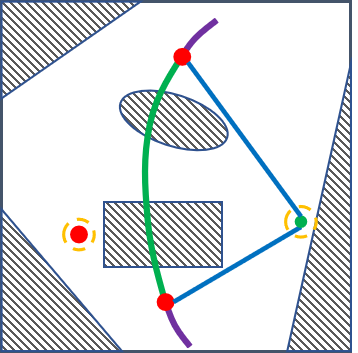}}       
		\subfigure[]
		{\includegraphics[width=0.4\columnwidth]{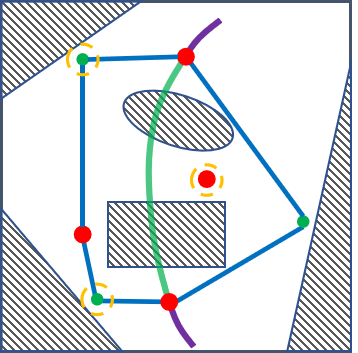}}       
		\subfigure[]
		{\includegraphics[width=0.4\columnwidth]{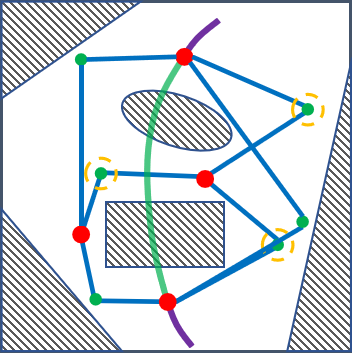}}       
		\subfigure[]
		{\includegraphics[width=0.4\columnwidth]{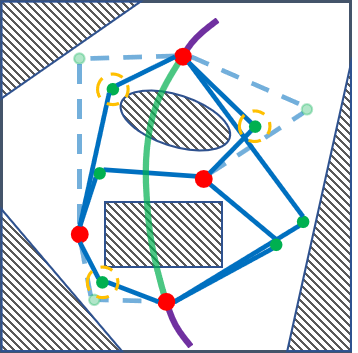}}       
		\subfigure[\label{fig:roadmap5}]
      {\includegraphics[width=0.4\columnwidth]{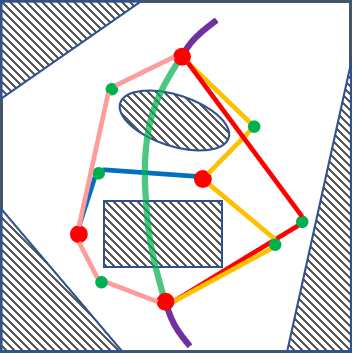}}       
      \vspace{-0.7cm}
	\end{center}
   \caption{\label{fig:roadmap} Generation of the topological roadmap. Red and green nodes represent the \textit{guards} and \textit{connectors} respectively.
   (a)-(c): \textit{Guards} are added to occupy different regions of space, and \textit{connectors} are added to form new connections between the \textit{guards}.
   (d): new \textit{connectors} replace the old ones, making the connections shorter. 
   (e): some of the paths found by the depth-first search. Both the red and orange paths belong to the same UVD class, while the pink path is the only member of its UVD class.}
   \vspace{-0.6cm}
\end{figure*}

Alg. \ref{alg:topo_roadmap} is used to construct a UVD roadmap $ \mathcal{G} $ capturing an abundant set of paths from different UVD classes.
Unlike standard PRM, which results in many redundant loops, our method generates a more compact roadmap where each UVD class contains just one or a few paths (displayed in Fig. \ref{fig:roadmap}).
% Therefore, only very short time is needed for pruning the redundant ones (Sect. \ref{subs:short_prune}).  

We introduce two different kinds of graph nodes, namely \textit{guard} and \textit{connector}, similar to the Visibility-PRM \cite{simeon2000visibility}.
The guards are responsible for exploring different part of the free space, and any two guards $ g_1 $ and $ g_2 $ are not \textit{visible} to each other (line $ \overline{g_1g_2} $ is in collision).
Before the main loop, two guards are created at the start point $ s $ and end point $ g $.
Every time a sampled point is invisible to all other guards, a new guard is created at this point (Line 6-7).
To form paths of the roadmap, connectors are used to connect different guards (Line 7-19).
When a sampled point is visible to exactly two guards, a new connector is created, either to connect the guards to form a topologically distinct connection (Line 19-20), or to replace an existing connector to make a shorter path (Line 16-17).
Limits of time ($t_{max}$) or sampling number ($ N_{max} $) are set to terminate the loop.

With the UVD roadmap, a depth-first search augmented by a visited node list is applied to search for the paths between $s$ and $g$, similar to \cite{rosmann2017integrated}.

% Before the main loop, two guards are created at the start and end points of the local trajectory (Line).
% To avoid redundant paths for each WVD class, the topology constraint is considered.
% Standard PRM method can result in many redundant paths for each WVD class, since the topology constraint is not considered.
% To reduce

\subsection{Path Shortening \& Pruning}
\label{subs:short_prune}
\begin{figure}[t]
	\begin{center}          
		\subfigure
		{\includegraphics[width=0.49\columnwidth]{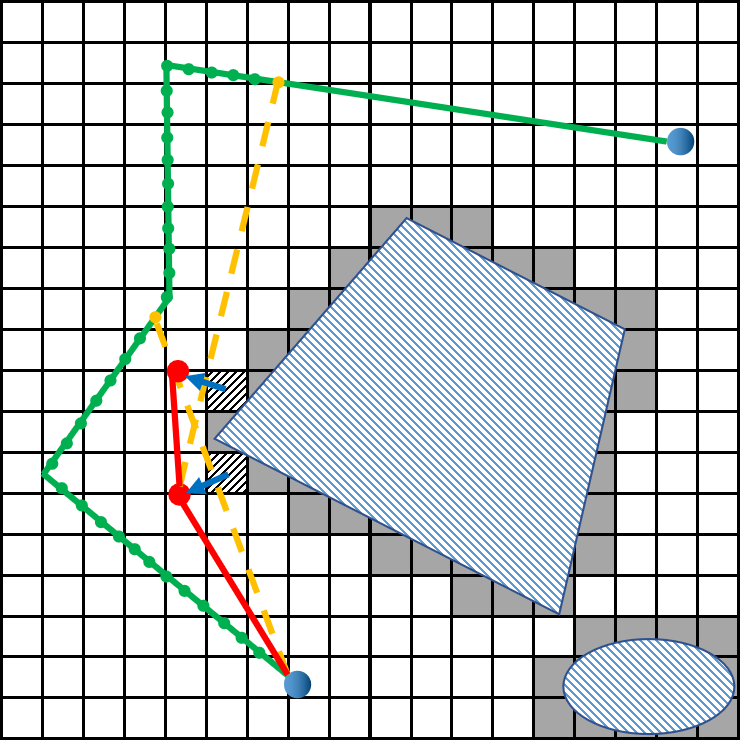}}       
		\subfigure
      {\includegraphics[width=0.49\columnwidth]{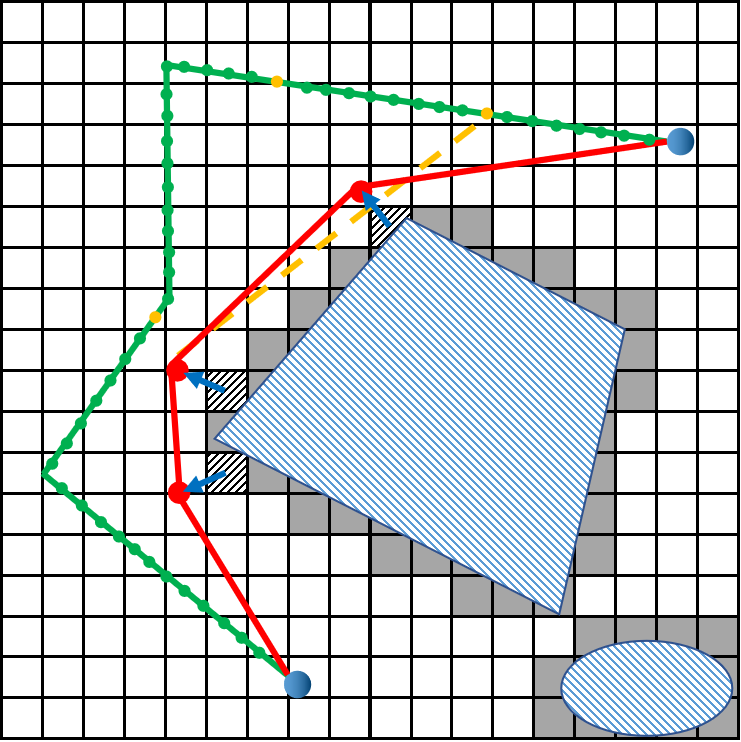}}       
      \vspace{-0.8cm}
	\end{center}
   \caption{\label{fig:shorten} A detoured and long path (green line) is shortened. 
   For each discretized point on the original path (green point), its visibility to the last waypoint of the shortened path (red points) is checked. 
   The center points of the blocking voxel (dashed) are pushed and appended as new waypoints.}
   \vspace{-0.9cm}
\end{figure} 
As shown in Fig. \ref{fig:roadmap5}, some paths obtained from Alg. \ref{alg:topo_roadmap} may be detoured. 
Such paths are unfavorable, since in the first phase of PGO it can deform a trajectory excessively and make it unsmooth. 
Hence, Alg. 2 find a topologically equivalent shortcut path $ \textit{P}_s $ for each $ \textit{P}_r $ found by the depth-first search (illustrated in Fig. \ref{fig:shorten}).
The algorithm uniformly \textbf{Discretize}s $ \textit{P}_r $ to a set of points $ \textit{P}_d $.
In each iteration, if a point $p_d$ in $ \textit{P}_d $ is invisible from the last point in $ \textit{P}_s $ (Line 3, 4), the center of the first occupied voxel blocking the view of $ \textit{P}_s.\textbf{back}() $ is found (Line 5).  
This point is then pushed away from obstacles in the direction orthogonal to $ l_d $ and coplanar to the ESDF gradient at $ p_b $ (Line 6), after which it is appended to $ \textit{P}_s $ (Line 7).
The process continues until the last point is reached.

Although in Alg. \ref{alg:topo_roadmap}, redundant connection between two guards are avoided, there may exist a small number of redundant paths between $s$ and $g$ (Fig. \ref{fig:roadmap5}).
To completely exclude repeated ones, we check the equivalence between any two paths and only preserve topologically distinct ones.

\begin{algorithm}
   \DontPrintSemicolon
   $ \textit{P}_d \gets \textnormal{\textbf{Discretize}}(\textit{P}_r), \ \textit{P}_s \gets \{\textit{P}_d.\textnormal{\textbf{front}}() \} $\;
   % $ \textit{P}_s.\textnormal{\textbf{push\_back}}() $\;
   \For{$ \textnormal{\textbf{each}} \ p_d \in \textit{P}_d $}{
      $ l_d \gets \textnormal{\textbf{Line}}(\textit{P}_s.\textnormal{\textbf{back}}(), p_d) $\;
      \If{$ \lnot \ \textnormal{\textbf{LineVisib}}(l_d) $}{
         $ p_b \gets \textnormal{\textbf{BlockPoint}}(l_d) $\;
         $ p_o \gets \textnormal{\textbf{PushAwayObs}}(p_b, l_d) $\;
         $ \textit{P}_s.\textnormal{\textbf{push\_back}}(p_o) $\;
         }
   }
   $ \textit{P}_s.\textnormal{\textbf{push\_back}}(\textit{P}_d.\textnormal{\textbf{back}}()) $\;
\caption{Finding a shortcut path $\textit{P}_s$ for $ \textit{P}_r $. \label{alg:search}}
\end{algorithm}

\section{Real-time Topological Trajectory Replanning}
\label{sec:topo_traj}
Algorithms in Sect. \ref{sec:topo_path} output a fruitful set of paths that can guide trajectory optimization.
We integrate them appropriately with the PGO for real-time replanning, as illustrated in Fig. \ref{fig:short_long}.
During the flight, a segment of the global trajectory within a specific horizon is checked periodically for safety.
Once collisions are detected, topological roadmap construction is triggered within a cube, which is determined by the start and end position of the segment, and $ (r_x, r_y, r_z) $ specifying the size of the cube.
Paths extracted from the roadmap are shortened and pruned (Sect. \ref{subs:short_prune}), after which each of the paths invokes an independent PGO.

Noticeably, the number of alternative UVD classes grows exponentially with the number of obstacles.
So in complex environments it may be intractable to optimize for all paths.
For this reason, we only select the first $ K_{max} $ shortest paths.
Paths more than $ r_{max} $ times longer than the shortest one are also excluded.
Such strategies bound the complexity and does not lead to the missing of potential optimality, since very long paths are unlikely to produce the optimal trajectory.
Practically we find $ K_{max} = 5 $ is sufficient.
% After a set of topologically distinct paths is found, shortened and pruned, the paths guide the PGO in parallel to generate a set of admissible trajectories (Fig. \ref{fig:topo_traj}). 
% Fortunately, in most cases we can bound the complexity tightly while not missing the potentially optimal solution.

\vspace{-0.1cm}
\section{Results}
\label{sec: result}

\subsection{Benchmark Comparisons}
\label{subs: benckmark}
\begin{figure}[t]
	\begin{center}          
		\subfigure
		{\includegraphics[width=0.457\columnwidth]{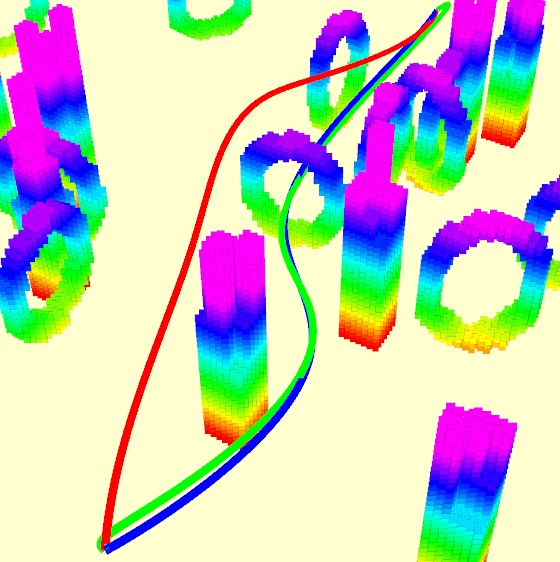}}       
      \vspace{-0.2cm}
		\subfigure
		{\includegraphics[width=0.45\columnwidth]{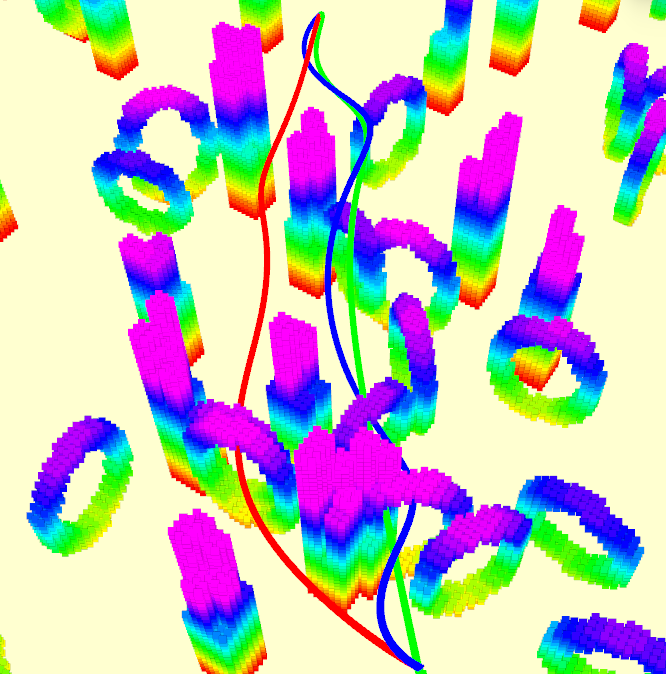}}       
		\subfigure
		{\includegraphics[width=0.9\columnwidth]{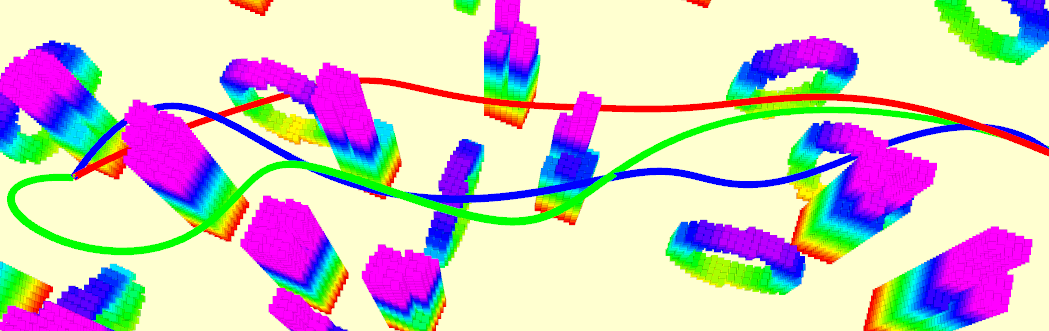}}       
      \vspace{-0.4cm}
	\end{center}
   \caption{\label{fig:benchmark} Benchmark comparisons of the proposed method (red) with Ewok\cite{usenko2017real} (green) and TRR\cite{gao2019teach} (blue) in environments of different complexities.}
   \vspace{-0.2cm}
\end{figure}

We first compare our local replanning method with two state-of-the-art methods Ewok\cite{usenko2017real} and TRR\cite{gao2019teach}.
Both methods parameterize local trajectory as uniform B-spline and use GTO to do replanning efficiently.
TRR further exploits the convex hull property of B-spline to simplify the cost function. 
We compare all methods in random maps with 3 different densities of obstacles, with 500 random replanning tasks for each density, as shown in Fig. \ref{fig:benchmark}.
For fair comparisons, we initialize all methods with the same reference trajectories computed by \cite{RicBryRoy1312}.
Besides, we limit the optimization time of Ewok and TRR to 15 $ ms $ according to their settings in the original work. 
For ours, the time for topological roadmap construction \footnote{The computation time of topological path searching can not be determined exactly, 
because time for path shortening and pruning varies slightly in different environments. 
So we determine the roadmap sampling time empirically according to the desired time budget.} and trajectory optimization is limited to 3 $ ms $ and 10 $ ms $ \footnote{Run-time efficiency is critical for online replanning, therefore we give short computation time to test the performance.}.
We check for collisions and smoothness of replanned trajectories.
% 10 random maps are generated for each density, and 50 replanning tasks with random start and end position are assigned in each map.

As is shown in Tab. \ref{tab:benchmark}, our method outperforms both benchmark methods in terms of success rate and smoothness.
Our method successfully finds feasible replanning in all environments, while the success rates of others decrease with increasing complexity of environments.
Also, our method generates the smoothest trajectories.
Our overall computation is only slightly longer, in which 5-7 $ ms $ is spent on topological path searching and $ 10 ms $ on parallel optimization.
It is notable that although less time is spent on optimization, the generated trajectories are better.
The reasons are that the proposed two-phases optimization (Sect. \ref{subs:pro_form}) is easier to converge,
and that the parallel optimization (Sect. \ref{sec:topo_traj}) explores the solution space more thoroughly.

% cccccc
\begin{table}
   \centering
   \caption{Comparisons of replanning methods. \label{tab:benchmark}}
   \begin{tabular}{ccccc} 
   \hline
   Density                  & Method  & \begin{tabular}[c]{@{}c@{}}Comp. \\time ($ms$) \end{tabular} & \begin{tabular}[c]{@{}c@{}}Success \\rate (\%)\end{tabular} & \begin{tabular}[c]{@{}c@{}}Smoothness\\($m^2/s^5$)\end{tabular}  \\ 
   \hline
   \hline
   \multirow{3}{*}{Low}     & Ewok\cite{usenko2017real}      & 15.00                                                         & 88.0                                                          & 9.0151                                                           \\ 
   \cline{2-5}
                             & TRR\cite{gao2019teach}        & 15.00                                                         & 90.4                                                          & 6.5102                                                           \\ 
   \cline{2-5}
                             & Proposed                      & 5.75 + 10.00                                                     & \textbf{100.0}                                                         & \textbf{5.4357}                                                           \\ 
   \hline
   \multirow{3}{*}{Medium}& Ewok\cite{usenko2017real}        & 15.00                                                         & 81.4                                                          & 9.5042                                                           \\ 
   \cline{2-5}
                             & TRR\cite{gao2019teach}        & 15.00                                                         & 85.6                                                          & 8.3942                                                           \\ 
   \cline{2-5}
                             & Proposed                      & 6.83 + 10.00                                                       & \textbf{100.0}                                                         & \textbf{6.7833}                                                           \\ 
   \hline
   \multirow{3}{*}{High} & Ewok\cite{usenko2017real}         & 15.00                                                         & 78.8                                                          & 9.4845                                                           \\ 
   \cline{2-5}
                             & TRR\cite{gao2019teach}        & 15.00                                                         & 82.6                                                          & 9.1762                                                           \\ 
   \cline{2-5}
                             & Proposed                      & 7.05 + 10.00                                                       & \textbf{100.0}                                                         & \textbf{7.7038}                                                           \\
   \hline
   \end{tabular}
   \vspace{-1.4cm}
   \end{table}

\subsection{Aggressive Autonomous Flights}
\label{subs:auto_flight}
We conducted aggressive autonomous flight experiments to validate the robustness of our replanning method.
The self-developed drone is localized by a robust visual-inertial state estimator \cite{qin2018vins}.
A local mapping framework \cite{han2019fiesta} fuses the depth images from a stereo camera into a volumetric occupancy map and maintains an ESDF for online replanning.
We use a geometric controller \cite{lee2010} for trajectory tracking.
All modules run on an Intel Core i7-8550U CPU.

\begin{figure}[t]
	\begin{center}          
		\subfigure
		{\includegraphics[width=0.50\columnwidth]{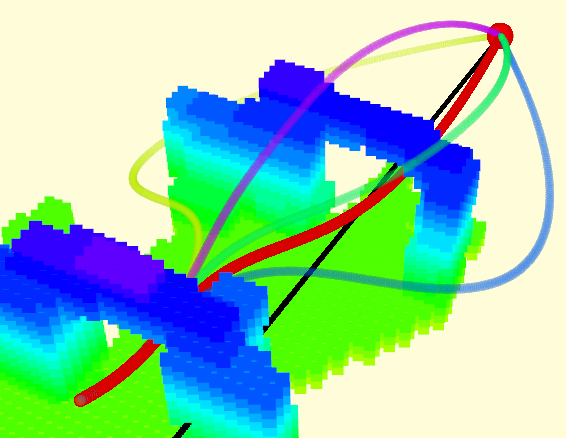}}       
		\subfigure
		{\includegraphics[width=0.48\columnwidth]{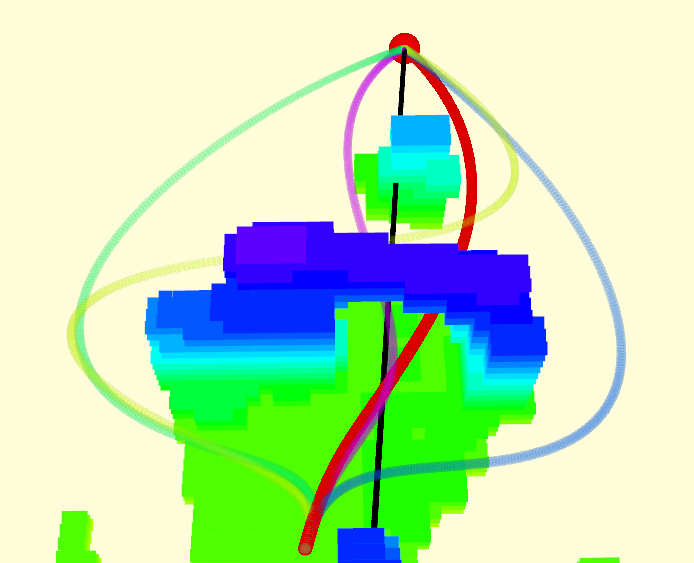}}       
		\subfigure
		{\includegraphics[width=0.7\columnwidth]{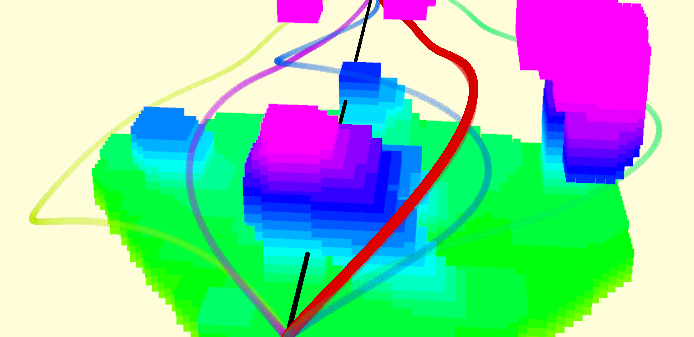}}       
      \vspace{-0.5cm}
	\end{center}
   \caption{\label{fig:flights} Autonomous flight experiments in indoor (top) and outdoor (buttom) environments. 
   A set of topologically distinct candidate trajectories are generated to avoid obstacles while keeping the drone close to the global reference trajectory (black). 
   The best one (red) is selected and executed.}
   \vspace{-0.4cm}
\end{figure}

The experiments are conducted in complex indoor and outdoor scenes.
In each experiment, a straight-line global reference trajectory is first generated using the approach \cite{RicBryRoy1312}.
During the flight, local trajectories within a horizon of 9 $m$ are replanned to avoid previously unknown obstacles while keeping the drone close to the global trajectory.
Aggressive autonomous flights with very limited sensing range in both scenes are quite challenging, since safe trajectories should be generated frequently within extremely short time to cope with sudden and unexpected obstacles.
The local trajectories, local maps and velocity profiles of one indoor and outdoor flights are shown in Fig. \ref{fig:flights}.
We refer the readers to the video attachment for more details.

\begin{figure}[t]
	\begin{center}          
		\subfigure
      {\includegraphics[width=0.483\columnwidth]{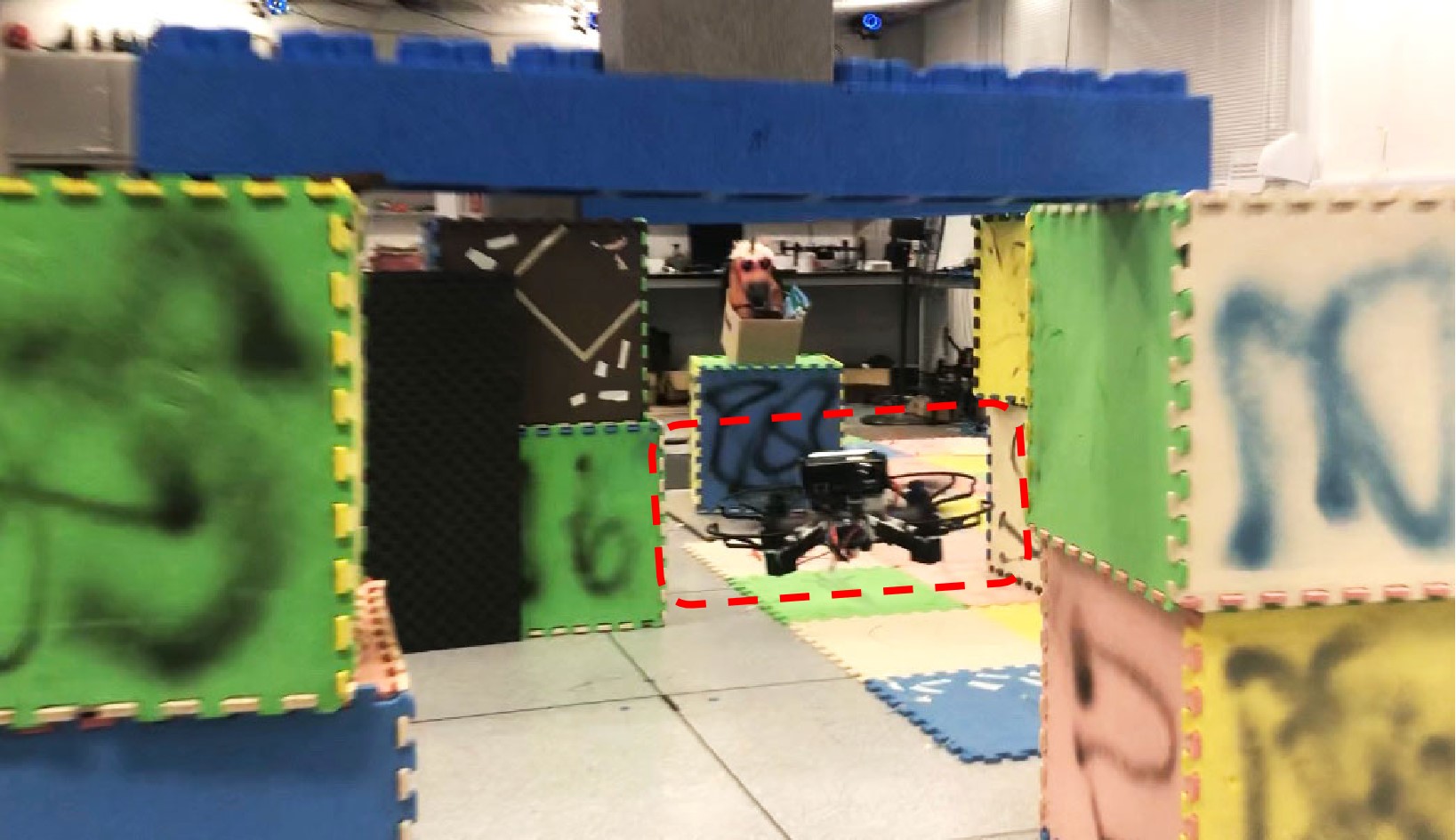}}       
		\subfigure
      {\includegraphics[width=0.5\columnwidth]{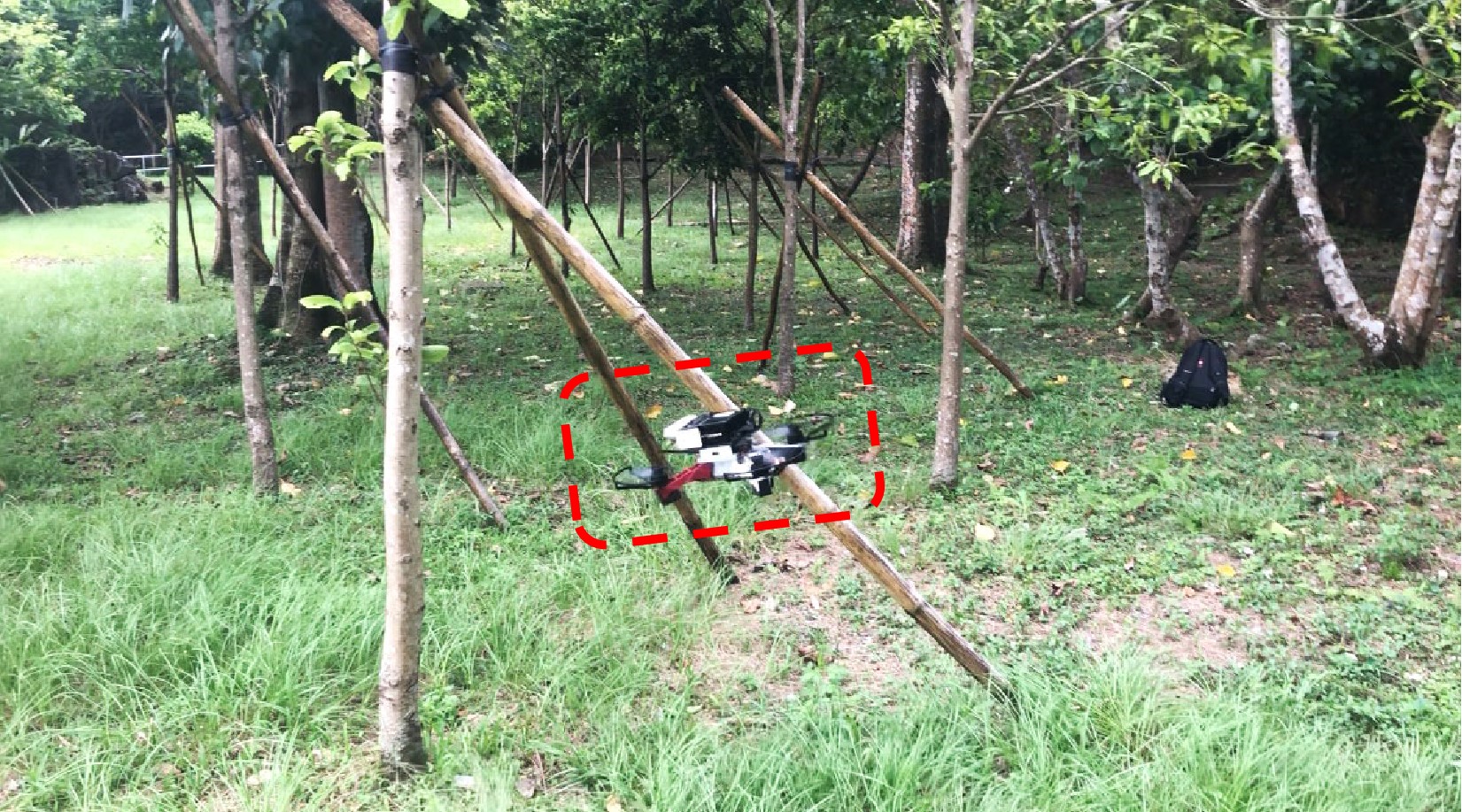}}
      \vspace{-0.8cm}       
	\end{center}
   \caption{\label{fig:velocity} Indoor and outdoor autonomous aggressive flights.}
   \vspace{-1.4cm}
\end{figure} 

\section{Conclusions}
\label{sec:conclude}
In this paper, we propose a robust trajectory replanning method for autonomous quadrotor navigation. 
It overcomes local minima with the path-guided optimization, topological path searching and parallel trajectory optimization.
Extensive benchmark comparisons and aggressive autonomous flight experiments are conducted to validate the robustness of our method.
% Benchmark comparisons show its superior success rate and optimality than state-of-the-art methods.
% Aggressive autonomous flight experiments in indoor and outdoor environments are presented to validate the robustness of our method. 

% A path-guided trajectory optimization is adopted to overcome infeasible local minima and it considerably improves the replanning success rate.
% A topological path searching is used to find multiple useful paths that capture the structures of 3-D environments and guide the trajectory optimization.
% It yields a more thorough exploration of the solution space and stronger optimality guarantee of the replanned trajectories.
% Benchmark comparisons show that our method is superior to state-of-the-art ones in both success rate and optimality aspects.

Currently, the performance of topological path searching is satisfactory, but its completeness is not analyzed in detail.
Also, we are not completely certain about the theoretical optimality of the replanning method.
In the future, we will investigate these problems.
We also plan to extend our method to cope with not only static but also dynamic obstacles to enable safe navigation in more complex scenes.

% Currently, our method does not plan the yaw trajectory to actively observe the unseen environment where potential collisions could happen.
% It only controls the camera of the drone to face the flying direcition, which can not guarantee safety. 
% In the future, we will develope a more effective strategy to ensure safety.  

\addtolength{\textheight}{-10.95cm}   % This command serves to balance the column lengths

\bibliography{zby}

\begin{thebibliography}{10}
\providecommand{\url}[1]{#1}
\csname url@samestyle\endcsname
\providecommand{\newblock}{\relax}
\providecommand{\bibinfo}[2]{#2}
\providecommand{\BIBentrySTDinterwordspacing}{\spaceskip=0pt\relax}
\providecommand{\BIBentryALTinterwordstretchfactor}{4}
\providecommand{\BIBentryALTinterwordspacing}{\spaceskip=\fontdimen2\font plus
\BIBentryALTinterwordstretchfactor\fontdimen3\font minus
  \fontdimen4\font\relax}
\providecommand{\BIBforeignlanguage}[2]{{%
\expandafter\ifx\csname l@#1\endcsname\relax
\typeout{** WARNING: IEEEtran.bst: No hyphenation pattern has been}%
\typeout{** loaded for the language `#1'. Using the pattern for}%
\typeout{** the default language instead.}%
\else
\language=\csname l@#1\endcsname
\fi
#2}}
\providecommand{\BIBdecl}{\relax}
\BIBdecl

\bibitem{oleynikova2016continuous}
H.~Oleynikova, M.~Burri, Z.~Taylor, J.~Nieto, R.~Siegwart, and E.~Galceran,
  ``Continuous-time trajectory optimization for online uav replanning,'' in
  \emph{Proc. of the {IEEE/RSJ} Intl. Conf. on Intell. Robots and
  Syst.({IROS})}, Daejeon, Korea, Oct. 2016, pp. 5332--5339.

\bibitem{fei2017iros}
F.~Gao, Y.~Lin, and S.~Shen, ``Gradient-based online safe trajectory generation
  for quadrotor flight in complex environments,'' in \emph{Proc. of the
  {IEEE/RSJ} Intl. Conf. on Intell. Robots and Syst.({IROS})}, Sept 2017, pp.
  3681--3688.

\bibitem{usenko2017real}
V.~Usenko, L.~von Stumberg, A.~Pangercic, and D.~Cremers, ``Real-time
  trajectory replanning for mavs using uniform b-splines and a 3d circular
  buffer,'' in \emph{2017 IEEE/RSJ International Conference on Intelligent
  Robots and Systems (IROS)}.\hskip 1em plus 0.5em minus 0.4em\relax IEEE,
  2017, pp. 215--222.

\bibitem{zhou2019robust}
B.~Zhou, F.~Gao, L.~Wang, C.~Liu, and S.~Shen, ``Robust and efficient quadrotor
  trajectory generation for fast autonomous flight,'' \emph{IEEE Robotics and
  Automation Letters}, vol.~4, no.~4, pp. 3529--3536, 2019.

\bibitem{gao2019teach}
F.~Gao, L.~Wang, B.~Zhou, L.~Han, J.~Pan, and S.~Shen, ``Teach-repeat-replan: A
  complete and robust system for aggressive flight in complex environments,''
  \emph{arXiv preprint arXiv:1907.00520}, 2019.

\bibitem{zucker2013chomp}
M.~Zucker, N.~Ratliff, A.~D. Dragan, M.~Pivtoraiko, M.~Klingensmith, C.~M.
  Dellin, J.~A. Bagnell, and S.~S. Srinivasa, ``Chomp: Covariant hamiltonian
  optimization for motion planning,'' \emph{The International Journal of
  Robotics Research}, vol.~32, no. 9-10, pp. 1164--1193, 2013.

\bibitem{kalakrishnan2011stomp}
M.~Kalakrishnan, S.~Chitta, E.~Theodorou, P.~Pastor, and S.~Schaal, ``Stomp:
  Stochastic trajectory optimization for motion planning,'' in \emph{2011 IEEE
  international conference on robotics and automation}.\hskip 1em plus 0.5em
  minus 0.4em\relax IEEE, 2011, pp. 4569--4574.

\bibitem{schmitzberger2002capture}
E.~Schmitzberger, J.-L. Bouchet, M.~Dufaut, D.~Wolf, and R.~Husson, ``Capture
  of homotopy classes with probabilistic road map,'' in \emph{IEEE/RSJ
  International Conference on Intelligent Robots and Systems}, vol.~3.\hskip
  1em plus 0.5em minus 0.4em\relax IEEE, 2002, pp. 2317--2322.

\bibitem{rosmann2015planning}
C.~R{\"o}smann, F.~Hoffmann, and T.~Bertram, ``Planning of multiple robot
  trajectories in distinctive topologies,'' in \emph{2015 European Conference
  on Mobile Robots (ECMR)}.\hskip 1em plus 0.5em minus 0.4em\relax IEEE, 2015,
  pp. 1--6.

\bibitem{rosmann2017integrated}
------, ``Integrated online trajectory planning and optimization in distinctive
  topologies,'' \emph{Robotics and Autonomous Systems}, vol.~88, pp. 142--153,
  2017.

\bibitem{bhattacharya2010search}
S.~Bhattacharya, ``Search-based path planning with homotopy class
  constraints,'' in \emph{Twenty-Fourth AAAI Conference on Artificial
  Intelligence}, 2010.

\bibitem{bhattacharya2012topological}
S.~Bhattacharya, M.~Likhachev, and V.~Kumar, ``Topological constraints in
  search-based robot path planning,'' \emph{Autonomous Robots}, vol.~33, no.~3,
  pp. 273--290, 2012.

\bibitem{jaillet2008path}
L.~Jaillet and T.~Sim{\'e}on, ``Path deformation roadmaps: Compact graphs with
  useful cycles for motion planning,'' \emph{The International Journal of
  Robotics Research}, vol.~27, no. 11-12, pp. 1175--1188, 2008.

\bibitem{oleynikova2018sparse}
H.~Oleynikova, Z.~Taylor, R.~Siegwart, and J.~Nieto, ``Sparse 3d topological
  graphs for micro-aerial vehicle planning,'' in \emph{2018 IEEE/RSJ
  International Conference on Intelligent Robots and Systems (IROS)}.\hskip 1em
  plus 0.5em minus 0.4em\relax IEEE, 2018, pp. 1--9.

\bibitem{blochliger2018topomap}
F.~Blochliger, M.~Fehr, M.~Dymczyk, T.~Schneider, and R.~Siegwart, ``Topomap:
  Topological mapping and navigation based on visual slam maps,'' in \emph{2018
  IEEE International Conference on Robotics and Automation (ICRA)}.\hskip 1em
  plus 0.5em minus 0.4em\relax IEEE, 2018, pp. 1--9.

\bibitem{schulman2014motion}
J.~Schulman, Y.~Duan, J.~Ho, A.~Lee, I.~Awwal, H.~Bradlow, J.~Pan, S.~Patil,
  K.~Goldberg, and P.~Abbeel, ``Motion planning with sequential convex
  optimization and convex collision checking,'' \emph{The International Journal
  of Robotics Research}, vol.~33, no.~9, pp. 1251--1270, 2014.

\bibitem{ding2018trajectory}
W.~Ding, W.~Gao, K.~Wang, and S.~Shen, ``Trajectory replanning for quadrotors
  using kinodynamic search and elastic optimization,'' in \emph{2018 IEEE
  International Conference on Robotics and Automation (ICRA)}.\hskip 1em plus
  0.5em minus 0.4em\relax IEEE, 2018, pp. 7595--7602.

\bibitem{liu2017iros}
S.~Liu, N.~Atanasov, K.~Mohta, and V.~Kumar, ``Search-based motion planning for
  quadrotors using linear quadratic minimum time control,'' in \emph{Proc. of
  the {IEEE/RSJ} Intl. Conf. on Intell. Robots and Syst.({IROS})}, Sept 2017,
  pp. 2872--2879.

\bibitem{simeon2000visibility}
T.~Sim{\'e}on, J.-P. Laumond, and C.~Nissoux, ``Visibility-based probabilistic
  roadmaps for motion planning,'' \emph{Advanced Robotics}, vol.~14, no.~6, pp.
  477--493, 2000.

\bibitem{RicBryRoy1312}
C.~Richter, A.~Bry, and N.~Roy, ``Polynomial trajectory planning for aggressive
  quadrotor flight in dense indoor environments,'' in \emph{Proc. of the Intl.
  Sym. of Robot. Research ({ISRR})}, Dec. 2013, pp. 649--666.

\bibitem{qin2017vins}
T.~Qin, P.~Li, and S.~Shen, ``Vins-mono: A robust and versatile monocular
  visual-inertial state estimator,'' \emph{arXiv preprint arXiv:1708.03852},
  2017.

\bibitem{han2019fiesta}
L.~Han, F.~Gao, B.~Zhou, and S.~Shen, ``Fiesta: Fast incremental euclidean
  distance fields for online motion planning of aerial robots,'' \emph{arXiv
  preprint arXiv:1903.02144}, 2019.

\bibitem{lee2010}
T.~Lee, M.~Leoky, and N.~H. McClamroch, ``Geometric tracking control of a
  quadrotor uav on se (3),'' in \emph{Proc. of the {IEEE} Control and Decision
  Conf. ({CDC})}, Atlanta, GA, Dec. 2010, pp. 5420--5425.

\end{thebibliography}

\end{document}